\documentclass[letterpaper, 10 pt, conference]{ieeeconf}  

\IEEEoverridecommandlockouts                              

\usepackage{graphics} 
\usepackage{epsfig} 
\usepackage{times} 
\usepackage{amsmath} 
\usepackage{amssymb}  

\usepackage{bm}
\usepackage{cite}
\usepackage{array}
\usepackage{xcolor}
\usepackage{textcomp}
\usepackage{amsfonts}
\usepackage{textcomp}
\usepackage[utf8]{inputenc} 
\usepackage[T1]{fontenc}    
\usepackage{hyperref}       
\usepackage{url}            
\usepackage{booktabs}       
\usepackage{nicefrac}       
\usepackage{microtype}      
\usepackage{xcolor}         
\usepackage{multirow}
\usepackage{algorithm}
\usepackage{algpseudocode}
\usepackage[caption=false]{subfig}

\graphicspath{{./fig/}}

\newcommand{\myPara}[1]{\vspace{.05in}\noindent\textbf{#1}}

\title{\LARGE \bf
Towards Personalized Federated Learning via Comprehensive Knowledge Distillation
}

\author{Pengju Wang, Bochao Liu, Weijia Guo, Yong Li, Shiming Ge
\thanks{Pengju Wang, Bochao Liu, Weijia Guo, Yong Li, and Shiming Ge are with the Institute of Information Engineering, Chinese Academy of Sciences, Beijing 100085, China, and also with the School of Cyber Security, University of Chinese Academy of Sciences, Beijing 100049, China. Emails: \{wangpengju, liubochao, guoweijia, liyong, geshiming\}@iie.ac.cn. Shiming Ge is the corresponding author. }
}

\begin{document}

\maketitle
\thispagestyle{empty}
\pagestyle{empty}

\begin{abstract}
Federated learning is a distributed machine learning paradigm designed to protect data privacy. However, data heterogeneity across various clients results in catastrophic forgetting, where the model rapidly forgets previous knowledge while acquiring new knowledge. To address this challenge, personalized federated learning has emerged to customize a personalized model for each client. However, the inherent limitation of this mechanism is its excessive focus on personalization, potentially hindering the generalization of those models. In this paper, we present a novel personalized federated learning method that uses global and historical models as teachers and the local model as the student to facilitate comprehensive knowledge distillation. The historical model represents the local model from the last round of client training, containing historical personalized knowledge, while the global model represents the aggregated model from the last round of server aggregation, containing global generalized knowledge. By applying knowledge distillation, we effectively transfer global generalized knowledge and historical personalized knowledge to the local model, thus mitigating catastrophic forgetting and enhancing the general performance of personalized models. Extensive experimental results demonstrate the significant advantages of our method.

\end{abstract}

\section{Introduction}

\begin{figure}[!t]
\centering 
    \subfloat[Data distribution]{%
        \includegraphics[width=0.496\linewidth]{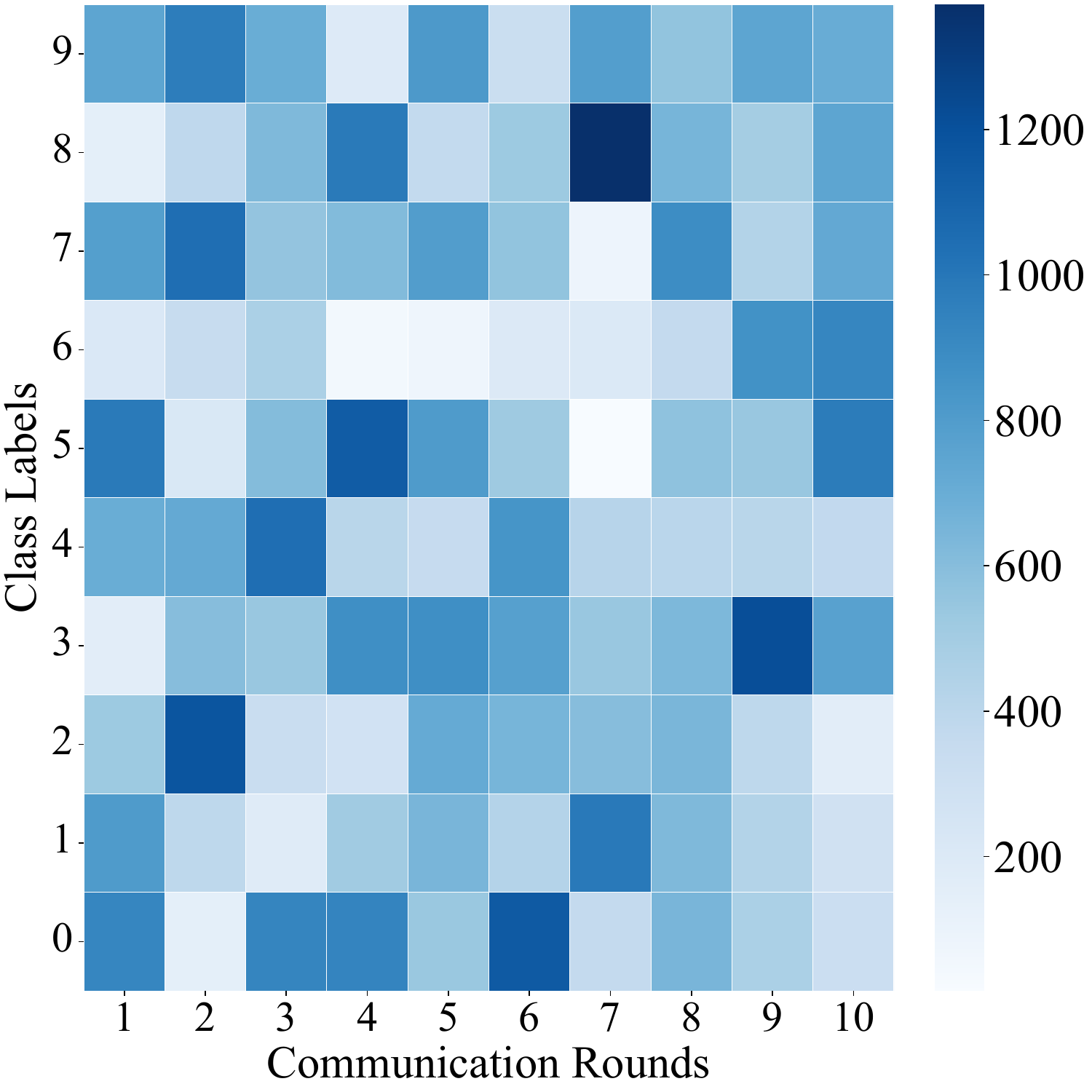}
        \label{fig:forgetting-a}
    }
    \subfloat[Catastrophic forgetting]{%
        \includegraphics[width=0.496\linewidth]{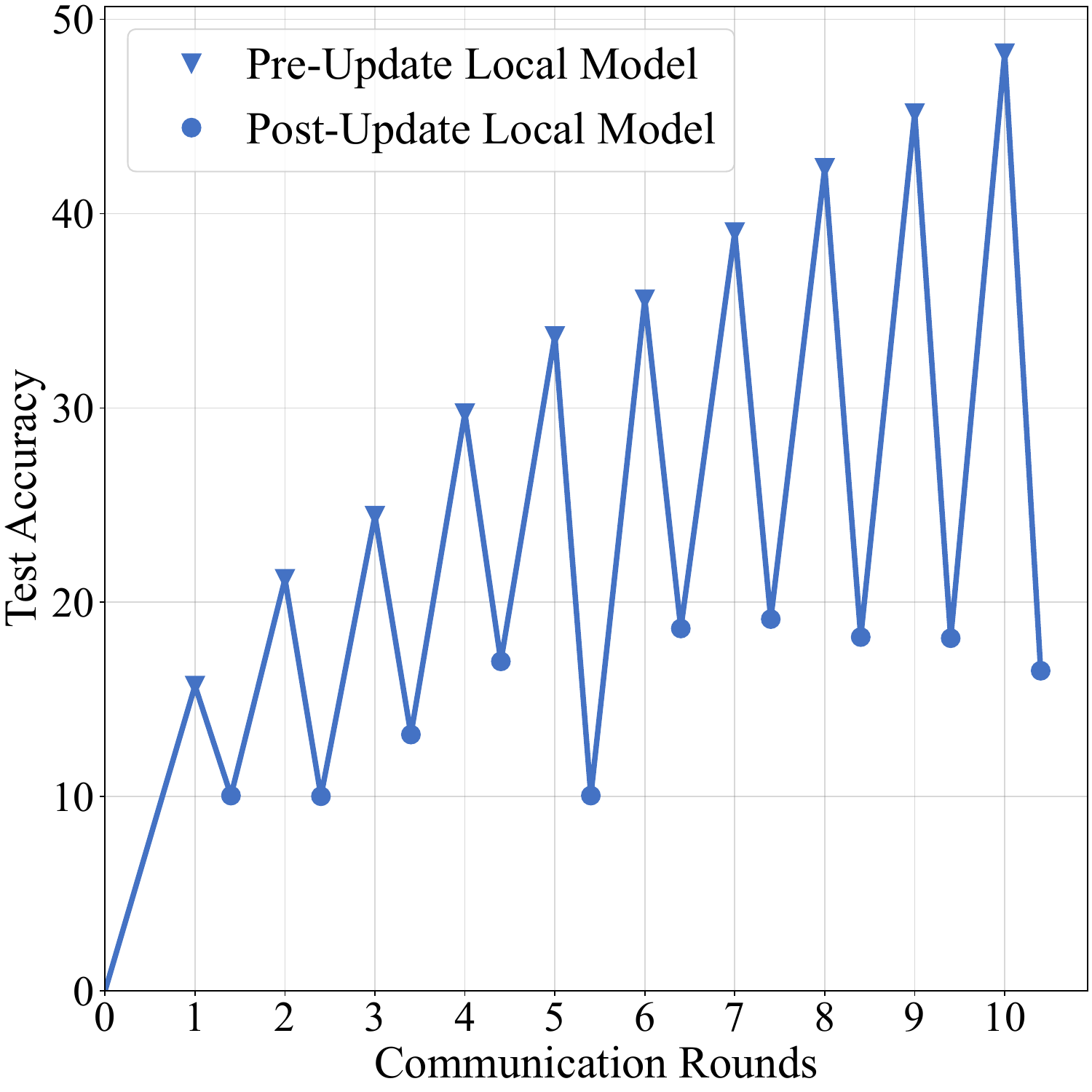}
        \label{fig:forgetting-b}
    }
    \caption{Data heterogeneity in FL leads to catastrophic forgetting.}
    \label{fig:catastrophic_forgetting}
\end{figure}

The rapid evolution of distributed intelligent systems has brought data privacy to the forefront. Federated Learning (FL), a distributed machine learning paradigm, enables collaborative model training through parameter sharing rather than raw data exchange, effectively reducing the risk of exposing sensitive data~\cite{mcmahan2017communication}. By leveraging FL, we can efficiently utilize data from distributed clients to collectively train high-performance models. FL has demonstrated its significant value in a variety of fields, including medical health~\cite{rana2024role}, financial analytics~\cite{abadi2024starlit}, and social network~\cite{chen2023dfedsn}. In traditional FL, multiple clients collaborate to train a global model to achieve an optimal universal solution for all clients. However, the non-independent and identically distributed nature of the data distribution~\cite{guo2023new}, known as data heterogeneity, often causes a decline in the performance of distributed clients and can even lead to catastrophic forgetting. Fig.~\ref{fig:catastrophic_forgetting} depicts this phenomenon. It is evident that the data distribution varies in each communication round. During each training round, distributed clients update their local model with the global model. Unfortunately, the post-update local model exhibits a significant performance decline compared to the pre-update local model, indicating the phenomenon of forgetting previously learned knowledge.

Personalized Federated Learning (PFL) is an innovative approach to tackle data heterogeneity in FL~\cite{tan2021towards}. It involves the collaborative training of a global model by all clients, after which each client develops a personalized model through personalized strategies, reflecting the distinctive characteristics of its local data. However, while existing PFL methods excel in model personalization, they often neglect model generalization. For instance, pFedSD~\cite{jin2023personalized} employs knowledge distillation to transfer knowledge from the historical model to the local model for achieving model personalization. Nevertheless, this method may hinder the local model's generalization, as the historical model represents the previous local model and incorporates personalized knowledge. The primary limitation of existing PFL methods lies in their design principles, which overemphasize personalized learning and may lead to model overfitting on individual clients, thereby reducing their adaptability to varied client~\cite{yi2024pfedmoe}.

In order to alleviate catastrophic forgetting and achieve a balance between generalization and personalization, we propose Personalized \underline{Fed}erated Learning via \underline{C}omprehensive \underline{K}nowledge \underline{D}istillation~(FedCKD). Our method integrates multi-teacher knowledge distillation~\cite{you2017learning} into FL for comprehensive knowledge transfer. The global model represents the aggregated model from the last round of server aggregation, containing global generalized knowledge, while the historical model represents the local model from the last round of client training, containing historical personalized knowledge. By employing the multi-teacher knowledge distillation, we utilize global and historical models as teachers and the local model as the student. This method effectively transfers both global generalized knowledge and historical personalized knowledge to the local model. Global generalized knowledge enhances model performance, whereas historical personalized knowledge addresses the issue of catastrophic forgetting.

In summary, our primary contributions are as follows: (1) We propose a novel PFL method called FedCKD. Through multi-teacher knowledge distillation, our method effectively transfers global generalized knowledge and historical personalized knowledge to the local model, thereby addressing catastrophic forgetting and enhancing model performance. (2) We introduce the annealing mechanism in knowledge distillation that dynamically adjusts the weight factor in the loss function, facilitating a smooth transition of training from knowledge transfer to local learning. This mechanism enhances the model's personalization ability and improves the training process's stability. (3) We validate the superior performance of our method through an extensive series of experiments, surpassing existing state-of-the-art methods.

\section{Related Work}

Federated Learning allows multiple clients to collaboratively train local models without sharing their data, addressing data privacy concerns. However, FL faces the challenge of data heterogeneity, which implies that data distribution across clients can vary significantly, potentially leading to subpar performance of traditional FL methods like FedAvg~\cite{mcmahan2017communication} and FedProx~\cite{li2020federated}. In response to this challenge, PFL has emerged. By introducing personalized strategies such as parameter decoupling, model regularization, personalized aggregation, or knowledge distillation, PFL aims to enhance the model's ability to learn from the unique data characteristics of each client while ensuring global generality. Through parameter decoupling, FedPer~\cite{arivazhagan2019federated} develops a global representation on the server while retaining a unique head for each client, LG-FedAvg~\cite{liang2020think} merges the benefits of localized representation learning with the training of a unified global model. Through model regularization, pFedMe~\cite{t2020personalized} leverages moreau envelopes as a regulatory loss to streamline local model training. Through personalized aggregation, FedFomo~\cite{zhang2020personalized} implements first-order model optimization in each client’s update phase to refine personalization. Through knowledge distillation, pFedSD~\cite{jin2023personalized} facilitates the transfer of personalized knowledge from the previous round of models to improve the current round. Although existing PFL methods have made significant progress, they frequently concentrate solely on enhancing the model personalization for individual clients, overlooking the vital aspect of maintaining model generalization. Future research should aim to achieve a balance between generalization and personalization, exploring novel strategies to enable models to offer personalized solutions for each client while maintaining adaptability to various clients.

\section{Methodology}

\subsection{Personalized Federated Learning}

In Traditional FL, training involves a server and $n$ clients. Each client, denoted as $C_k$ ($k=1,2,\ldots,n$), possesses its own private data $\mathcal{D}_{k}=\{\bm{x}_{k}, \bm{y}_{k}\}$, with $\bm{x}_{k}$ denotes the sample and $\bm{y}_{k}$ denotes the label. The collective data across all clients is represented as $\mathcal{D}={\bigcup_{k=1}^{n}}\mathcal{D}_{k}$. The goal in FL is to derive a global model $\bm{w}_{g}$ that minimizes the overall loss function,
\begin{equation}
    \min_{\bm{w}_{g}} \sum_{k=1}^{n}\frac{\left| \mathcal{D}_k \right |}{\left | \mathcal{D} \right |} \mathcal{L}_{k}(\mathcal{D}_k;\bm{w}_{g}),
\end{equation}
where $\mathcal{L}_{k}(\mathcal{D}_k;\bm{w}_{g})$ is the empirical loss for client $C_k$.

The training process comprises critical phases. In the server broadcast phase, the server sends its global model $\bm{w}_{g}$ to all clients. In the client update phase, each client replaces its local model, denoted as $\bm{w}_{k} \leftarrow \bm{w}_{g}$, and conducts local updates by applying $\bm{w}_{k} \leftarrow \bm{w}_{k} - \eta \nabla \mathcal{L}_{k}(\mathcal{D}_{k}; \bm{w}_{k})$ to train its local model with own data. In the client upload phase, each client uploads its updated local model $\bm{w}_{k}$ back to the server. In the global aggregation phase, the server aggregates the local models to generate a global model $\bm{w}_{g} \leftarrow \sum_{k=1}^{n}\frac{\left| \mathcal{D}_k \right |}{\left | \mathcal{D} \right |} \bm{w}_{k}$. These above phases are iteratively executed until the model converges or a predefined number of training rounds is reached.

\begin{figure*}[t]
\centering
\includegraphics[width=0.76\linewidth]{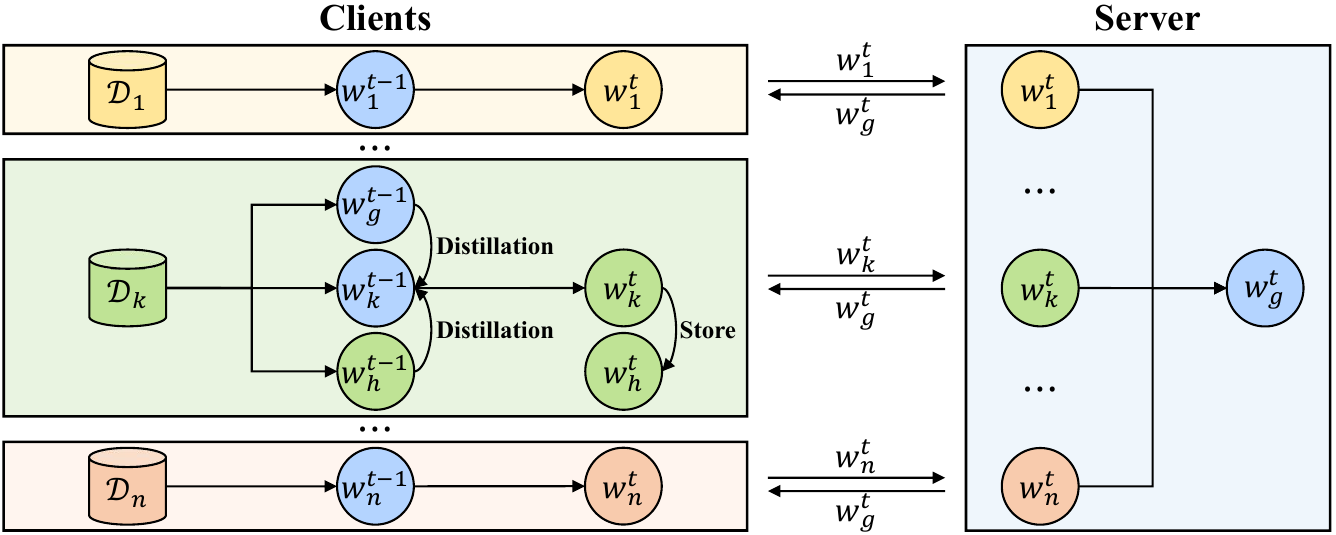} 
\caption{The framework of our method. The client update phase is divided into local distillation and local store processes. During the local distillation process, a comprehensive knowledge distillation is performed, transferring knowledge from the global model $\bm{w}^{t-1}_g$ and the historical model $\bm{w}^{t-1}_h$ to the local model $\bm{w}^{t-1}_k$. During the local store process, the emphasis is on preserving the local model for the next round, $\bm{w}^{t}_h\leftarrow\bm{w}^{t}_k$. Different colors are used to differentiate between knowledge types, where blue represents generalized knowledge and other colors indicate personalized knowledge.} 
\label{fig:method_framework} 
\end{figure*}

Despite its clear advantages in overall performance, the global model $\bm{w}_{g}$ suffers a significant performance decline in heterogeneous data scenarios. This decline is largely due to the global model's objective in FL to be universally applicable, which often does not cater to the unique data of individual clients. To address this, PFL customizes a personalized model $\bm{w}_k$ for each client, effectively mitigating the challenges of data heterogeneity and enabling a tailored exploration of each client’s unique data characteristics. The optimization problem in PFL is formulated as follows:
\begin{equation}
    \min_{\bm{w}_1,\dots,\bm{w}_n} \sum_{k=1}^{n}\frac{\left| \mathcal{D}_k \right |}{\left | \mathcal{D} \right |} \mathcal{L}_{k}(\mathcal{D}_k; \bm{w}_k).
\end{equation}

\textit{Should we focus on the model's generalized performance, representing its performance across extensive data, or on its personalized performance, representing its adaptability to specific data?} Striking a balance between these two aspects is essential in crafting personalized strategies that effectively capture the unique data distribution characteristics of each client while maintaining generality.

\subsection{Comprehensive Knowledge Distillation}

Knowledge distillation (KD) transfers knowledge from a trained teacher model to an untrained student model~\cite{hinton2015distilling}. Let $p_t$ and $p_s$ denote the outputs of the teacher and student models, respectively. To smooth the output distribution, a temperature parameter $\tau$ is used in the softmax function $p^\tau=\frac{\exp(z_i/\tau)}{\sum_j \exp(z_j/\tau)}$, where $z_i$ is the $i$-th output element. KD aims to minimize the discrepancy between the teacher and student models. The loss function is defined as follows:
\begin{equation}
    \mathcal{L}=\mathcal{L}_{\rm CE}(p_s, y) + \lambda  \mathcal{L}_{\rm KL}(p_s^\tau, p_t^\tau),
\end{equation}
where $\mathcal{L}_{\rm CE}$ is the Cross-Entropy loss between the soft labels $p_s$  and the hard labels $y$, $\mathcal{L}_{\rm KL}$ is the Kullback-Leibler divergence loss between the soft labels $p_s^\tau$ and the soft labels $p_t^\tau$, $\lambda$ is the weight factor.

Recently, KD has become a critical technique in FL~\cite{qin2024knowledge}, offering innovative solutions to tackle the challenge of catastrophic forgetting, exemplified by methods such as pFedSD~\cite{jin2023personalized}. This method facilitates the transfer of knowledge from the historical model to the local model for each client. However, it places excessive emphasis on utilizing personalized knowledge from the historical model while overlooking generalized knowledge of the global model.

As shown in Fig.~\ref{fig:method_framework}, We propose a comprehensive knowledge distillation method using multi-teacher knowledge distillation. In our method, we retained the local model from previous training, referred to as the historical model, to maintain personalized knowledge. The global model, in contrast, embodies generalized knowledge. We employ both the global and historical models as teachers, with the local model serving as the student, to facilitate a thorough knowledge transfer that balances generalization and personalization. The loss function is defined as follows:
\begin{equation}
    \mathcal{L}_k=\mathcal{L}_{\rm CE}(p_k, y) + \lambda  \mathcal{L}_{\rm KL}(p_k^\tau, p_g^\tau) + \lambda \mathcal{L}_{\rm KL}(p_k^\tau, p_h^\tau),
\end{equation}
where $p_k^\tau$ represents the soft labels of the local model $\bm{w}_{k}$, $p_g^\tau$ represents the soft labels of the global model $\bm{w}_{g}$, and $p_h^\tau$ represents the soft labels of the historical model $\bm{w}_{h}$.

To facilitate effective knowledge transfer, we implement an annealing mechanism. Initially, the student model primarily learns from the teacher model's soft labels. As training advances, $\lambda$ undergoes annealing, gradually increasing its emphasis on the hard labels. We employ exponential decay for annealing, which simplifies the process by gradually decreasing the weight factor $\lambda$ at a constant decay rate $\gamma$ for each communication round.

\begin{algorithm}[t]
    \caption{FedCKD}
    \label{algorithm:FedCKD}
    \begin{algorithmic}[1]
        \Require Number of clients $n$, participation rate $r$, communication rounds $T$, local epochs $E$, learning rate $\eta$.
        \Ensure Personalized models $\bm{w}_{1}, \bm{w}_{2}, \ldots, \bm{w}_{n}$
        \Procedure{Server}{}
        \State Server initializes $\bm{w}^{0}_{g}$
            \For{round $t=1,2,\ldots,T$}
                \State Server samples clients $K \leftarrow rn$
                \State Server sends $\bm{w}^{t-1}_{g}$ to clients $K$
                \For{client $k=1,2,\ldots,K$}
                    
                    \For{epoch $e = 1,2,\ldots,E$}
                        \State $\bm{w}^{t}_{k}\leftarrow$ \textsc{Client}$(\mathcal{D}_{k}; \bm{w}^{t-1}_{k}, \bm{w}^{t-1}_{g}, \bm{w}^{t-1}_{h})$
                    \EndFor 
                \EndFor
                \State Server aggregates $\bm{w}^{t}_{g} \leftarrow \sum_{k=1}^{K}\frac{\left| \mathcal{D}_k \right |}{\left | \mathcal{D} \right |} \bm{w}^{t}_{k}$
            \EndFor
            \State \Return $\bm{w}_{1}, \bm{w}_{2}, \ldots, \bm{w}_{n}$
        \EndProcedure
        \Procedure{Client}{$\mathcal{D}_{k}; \bm{w}^{t-1}_{k}, \bm{w}^{t-1}_{g}, \bm{w}^{t-1}_{h}$}
                \State Client updates $\bm{w}^{t-1}_{k}\leftarrow\bm{w}^{t-1}_{g}$
                \For{local data $\mathcal{D}_{k} $}
                    \State $\bm{w}^{t}_{k} \leftarrow \bm{w}^{t-1}_{k} - \eta \nabla \mathcal{L}_{k}(\mathcal{D}_{k}; \bm{w}^{t-1}_{k}, \bm{w}^{t-1}_{g}, \bm{w}^{t-1}_{h})$
                \EndFor
                \State Client stores $\bm{w}^{t}_{h}\leftarrow\bm{w}^{t}_{k}$
            \State \Return $\bm{w}^{t}_{k}$ to server
        \EndProcedure
    \end{algorithmic}
\end{algorithm}

Algorithm~\ref{algorithm:FedCKD} shows FedCKD in detail. During each round, the server selects a subset of clients at random and sends the global model to them. These clients subsequently update their local model with the global model. Comprehensive knowledge distillation is then performed to transfer knowledge from both the global and historical models to the local model, and the trained local model is preserved as the historical model for future training rounds. Lastly, clients send their local model back to the server, where multiple local models are aggregated to create an enhanced global model.

\begin{figure}[t]
\centering 
    \subfloat[Practical setting $\alpha=0.10$]{%
        \includegraphics[width=0.496\linewidth]{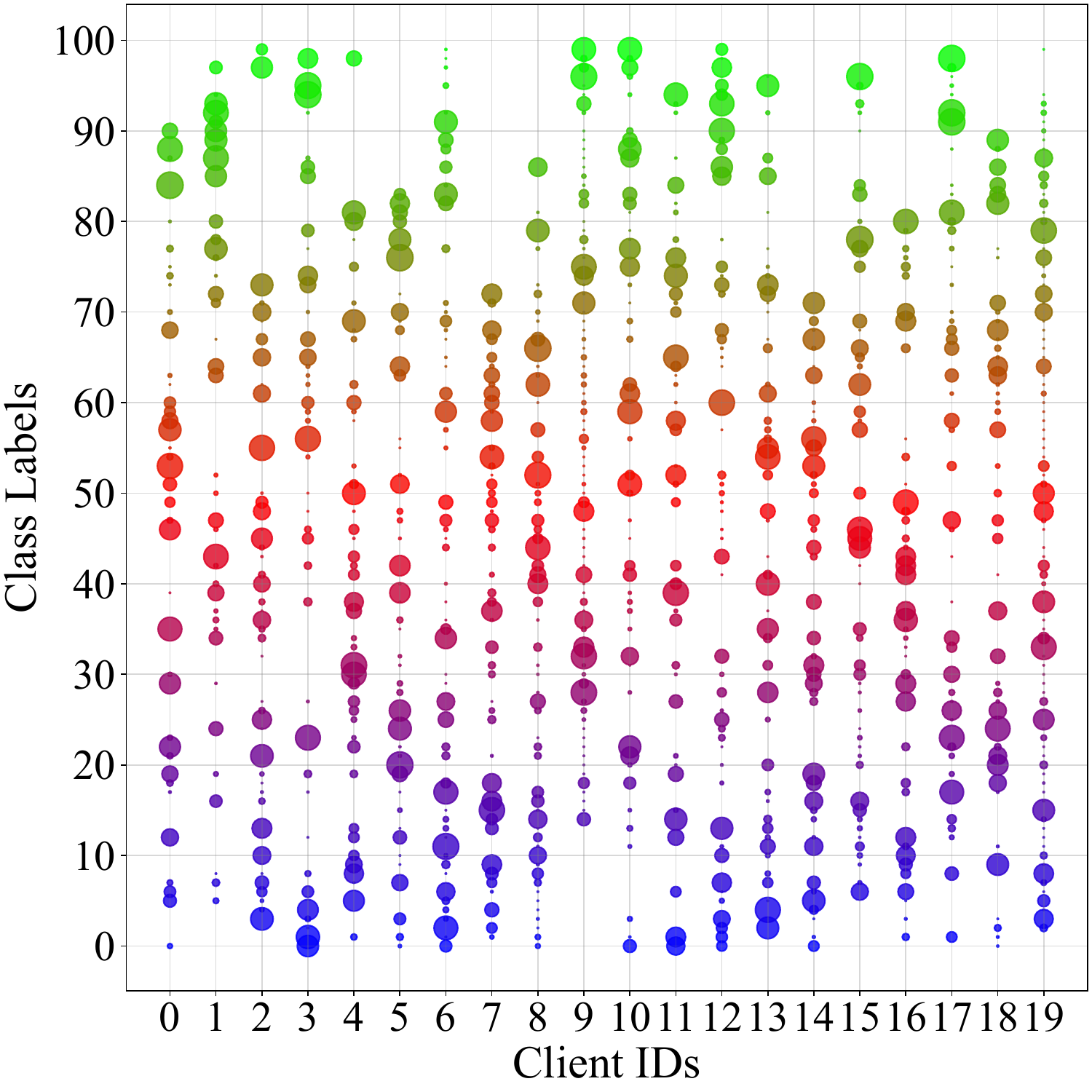}
        \label{fig:heterogeneity_a}
    }
    \subfloat[Pathological setting $s=20$]{%
        \includegraphics[width=0.496\linewidth]{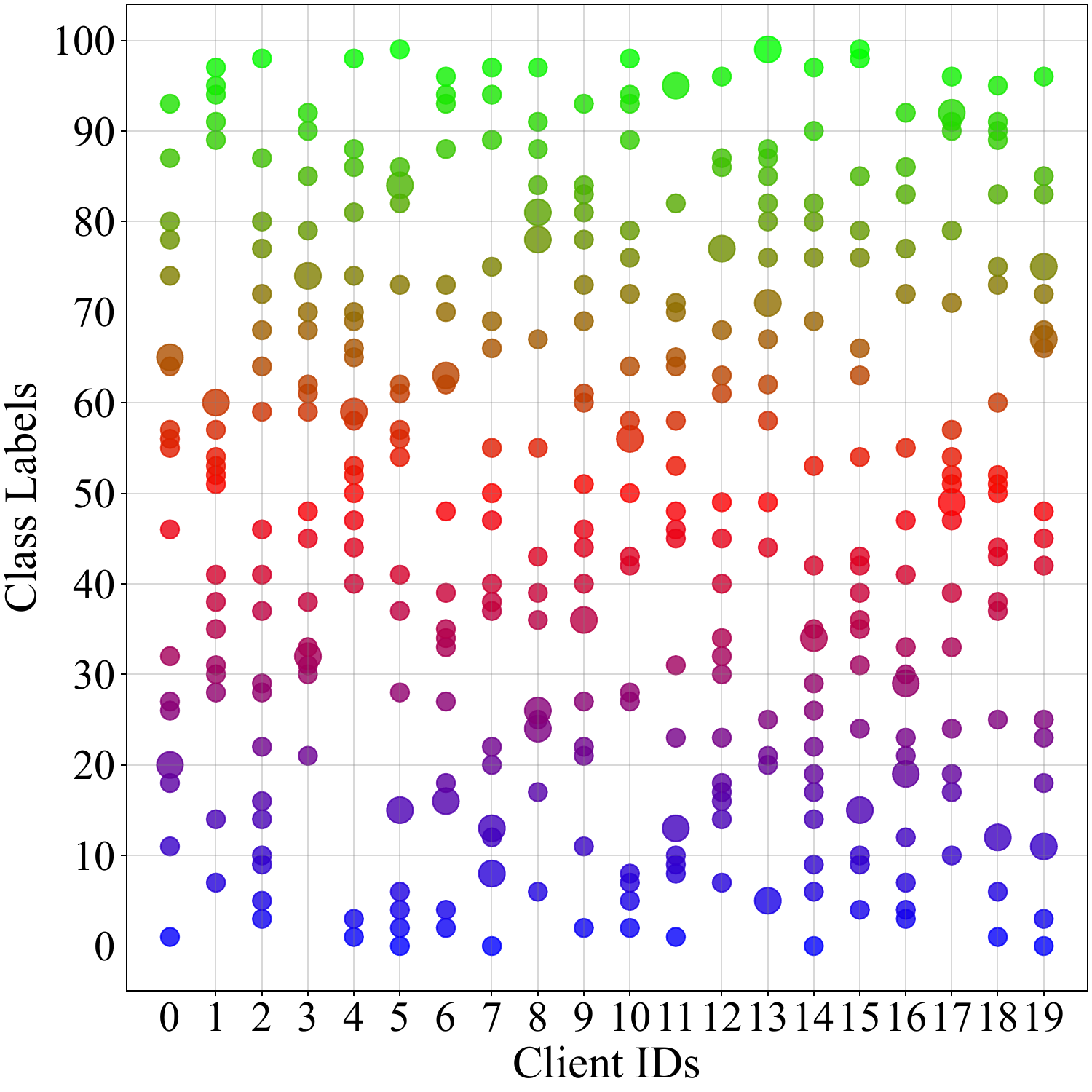}
        \label{fig:heterogeneity_b}
    }
    \caption{Data heterogeneity among $20$ clients on the CIFAR100 dataset.}
    \label{fig:data_heterogeneity}
\end{figure}

\section{Experiments}

\subsection{Experimental Setup}

\myPara{Datasets.}~We simulate two heterogeneous settings using three datasets: FMNIST~\cite{xiao2017fashion}, CIFAR10~\cite{krizhevsky2009learning}, and CIFAR100~\cite{krizhevsky2009learning}. In the practical setting, data distribution reflects real-world diversity, characterized by a Dirichlet distribution $Dir(\alpha)$, with a default $\alpha=0.10$ for all datasets. In the pathological setting, data distribution is extremely unbalanced, where each client is assigned imbalanced data from $s=2/2/20$ classes out of the $10/10/100$ classes in the FMNIST/CIFAR10/CIFAR100 datasets. Fig.~\ref{fig:data_heterogeneity} illustrates the distribution of the CIFAR100 dataset among $20$ clients, showcasing sample numbers by circle size and sample labels by circle color.

\myPara{Models.}~We utilize two commonly used models: a simple CNN for the FMNIST dataset, and a five-layer CNN for the CIFAR10 and CIFAR100 datasets, adhering to the model outlined in \cite{jin2023personalized} to guarantee fairness.

\myPara{Baselines.}~We contrast our method FedCKD with state-of-the-art methods. For traditional FL methods, we evaluate against FedAvg~\cite{mcmahan2017communication} and FedProx~\cite{li2020federated}. For PFL methods, we evaluate against FedPer~\cite{arivazhagan2019federated}, LG-FedAvg~\cite{liang2020think}, pFedMe~\cite{t2020personalized}, FedFomo~\cite{zhang2020personalized}, and pFedSD~\cite{jin2023personalized}. 

\myPara{Hyperparameters.}~We explore two scenarios: (1) $n=20$ clients with a participation rate of $r=1.00$ over $T=50$ communication rounds; (2) $n=100$ clients with a participation rate of $r=0.10$ over $T=100$ communication rounds. The training process involves executing local epochs $E=5$ and batch sizes $B=64$. The SGD optimizer is used with a learning rate of $0.01$, a momentum factor of $0.90$, and a weight decay of $1e-5$. For comparability, our method sets the weight factor to $\lambda=0.50$ and distillation temperature to $\tau=3$ in line with \cite{jin2023personalized}, while the decay rate for the annealing mechanism is set to $\gamma=0.99$. When employing alternative methods, we strictly adhere to the settings detailed in \cite{jin2023personalized}. We repeat each experiment three times, with the recording of both mean and standard deviation. Evaluations are conducted based on the average test accuracy across all clients.

\begin{figure}[t]
\centering 
    \subfloat[Client scale $n=20$]{%
        \includegraphics[width=0.496\linewidth]{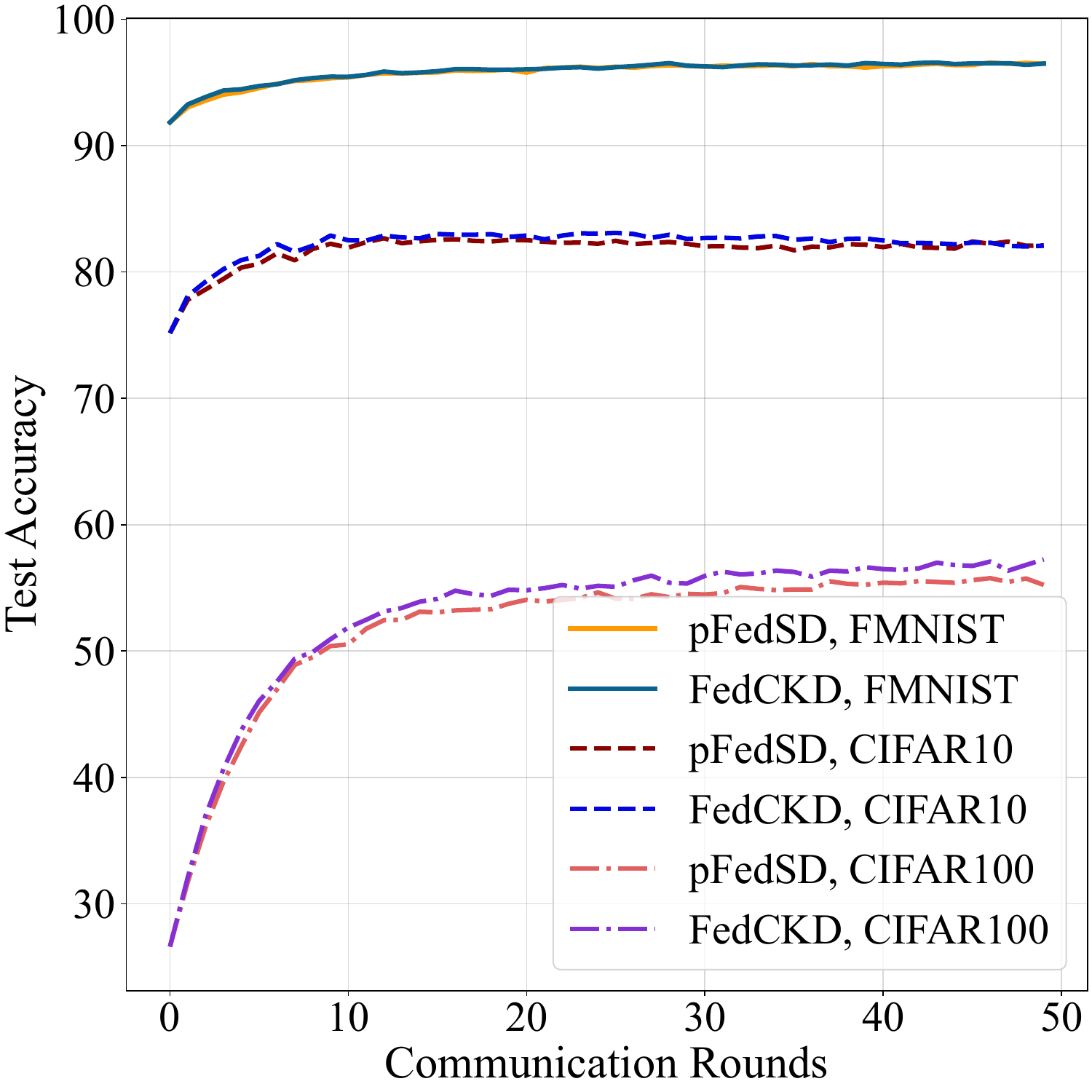}
        \label{fig:round-a}
    }
    \subfloat[Client scale $n=100$]{%
        \includegraphics[width=0.496\linewidth]{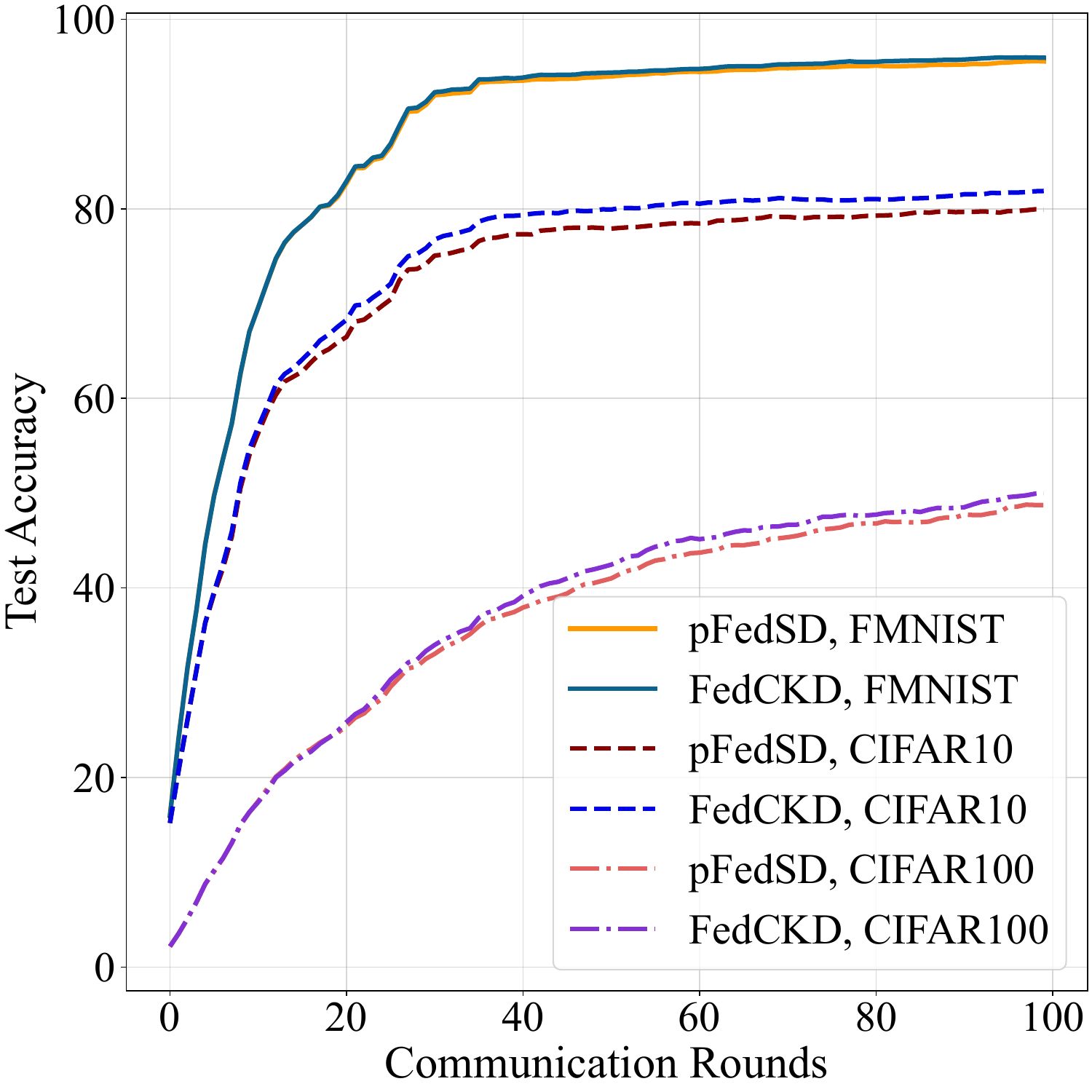}
        \label{fig:round-b}
    }
    \caption{Learning curves under different experimental settings.}
    \label{fig:main_convergence}
\end{figure}

\begin{figure}[t]
\centering 
    \subfloat[Practical setting $\alpha=0.10$]{%
        \includegraphics[width=0.496\linewidth]{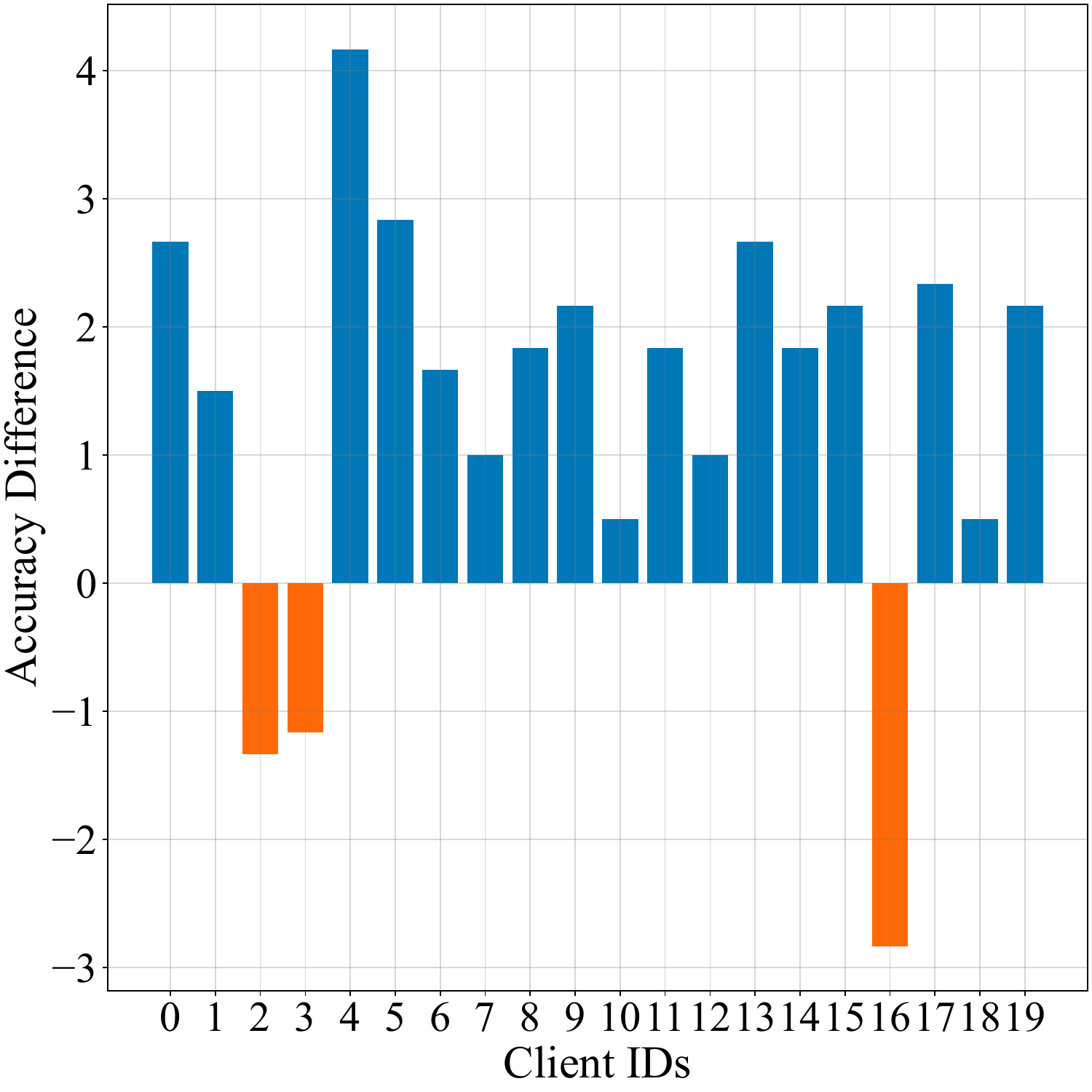}
        \label{fig:difference-a}
    }
    \subfloat[Pathological setting $s=20$]{%
        \includegraphics[width=0.496\linewidth]{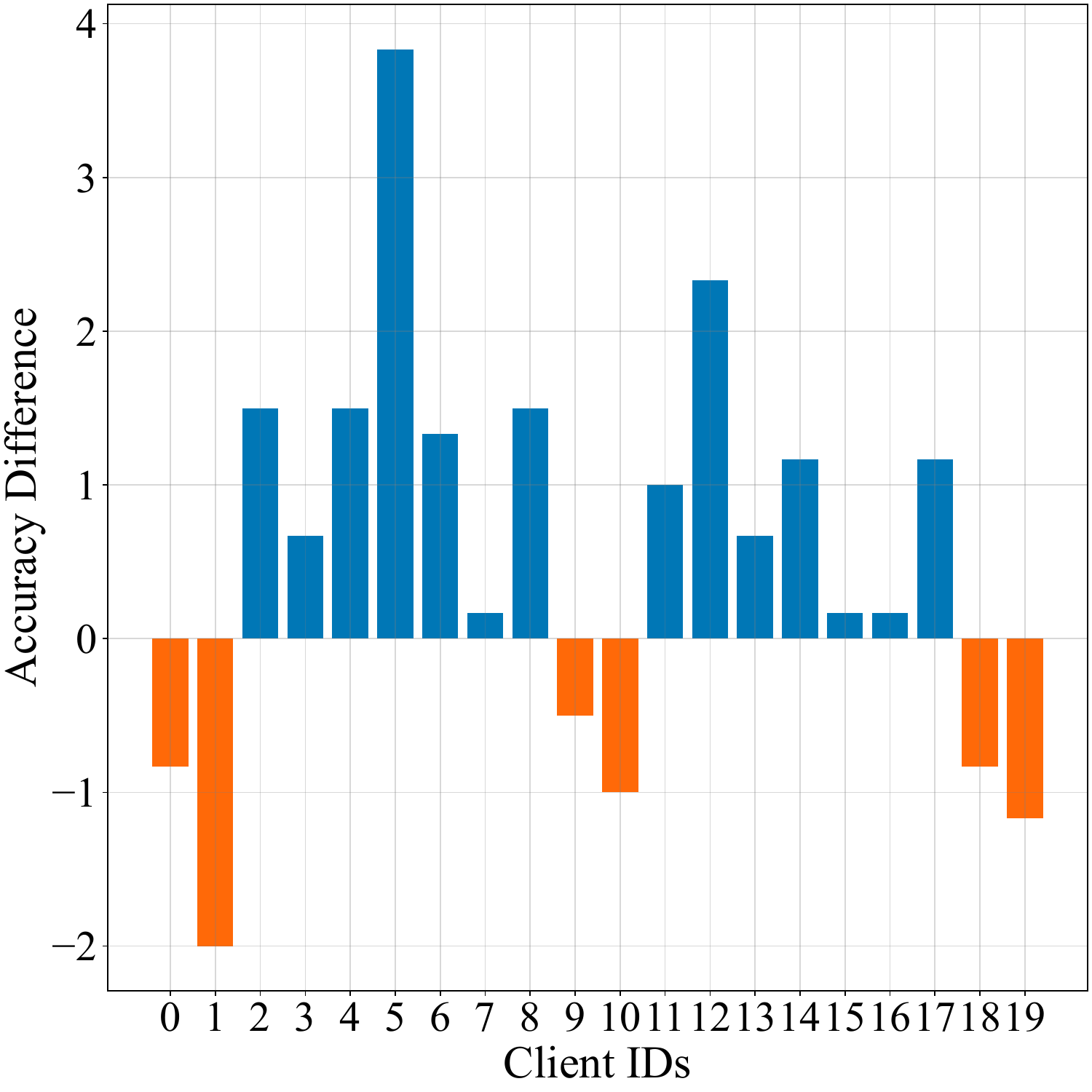}
        \label{fig:difference-b}
    }
    \caption{Accuracy difference~(\%) among $20$ clients on the CIFAR100 dataset.}
    \label{fig:accuracy_difference}
\end{figure}

\subsection{Performance Comparison}

\myPara{Test accuracy.}~We assess the test accuracy across three datasets in the practical and pathological settings, as presented in Table~\ref{tab:main_practical}--~\ref{tab:main_pathological}. Our method, FedCKD, consistently outperforms other methods in various settings. A notable highlight is its performance in the practical setting on the CIFAR100 dataset with $20$ clients, where FedCKD exceeded the closest baseline, pFedSD, by a significant margin of $1.98\%$. Furthermore, FedCKD exhibits remarkable stability, as indicated by its low standard deviation in most cases. On the contrary, the performance of FedAvg and FedProx is subpar because they fail to adequately account for the diverse distribution of data. FedPer and LG-FedAvg solely share partial parameters of the local model, overlooking vital local knowledge embedded in other parameters. pFedMe enhances the model's alignment with each client's data by incorporating regularization terms. Nevertheless, an excessive emphasis on personalization could compromise the global model's generalization prowess. FedFomo relies on client-specific weights for model aggregation, potentially diminishing the cohesive integration of global knowledge. pFedSD leverages personalized knowledge from historical models to enhance performance, yet overlooks generalized knowledge within the global model. The learning curve illustrates how a model's accuracy evolves with the increasing number of communication rounds. In the practical setting, as shown in Fig.~\ref{fig:main_convergence}, FedCKD consistently outperforms pFedSD across all datasets and communication rounds, affirming the effectiveness of FedCKD. Overall, FedCKD significantly improves model accuracy in heterogeneous settings by balancing personalized and generalized knowledge.

\begin{table*}[!htb]
    \centering
    \caption{Test accuracy~(\%) on different datasets in the practical setting}
    \label{tab:main_practical}
    \resizebox{1.0\linewidth}{!}{
        \begin{tabular}{llcccccccc}
        \toprule
        Dataset & Client & FedAvg & FedProx & FedPer & LG-FedAvg & pFedMe & FedFomo & pFedSD & FedCKD  \\ 
        \midrule
        \multirow{2}{*}{FMNIST} 
        & $n=20$ & 90.15$\pm$0.10 & 90.05$\pm$0.13 & 96.30$\pm$0.03 & 95.22$\pm$0.22 & 93.24$\pm$0.08 & 95.38$\pm$0.04 & 96.57$\pm$0.08 & {\bf96.61$\pm$0.06} \\
        & $n=100$ & 86.82$\pm$0.39 & 76.78$\pm$0.30 & 94.99$\pm$0.17 & 91.41$\pm$0.08 & 92.05$\pm$0.08 & 92.06$\pm$0.14 & 95.97$\pm$0.06 & {\bf95.98$\pm$0.04} \\
        \midrule
        \multirow{2}{*}{CIFAR10} 
        & $n=20$ & 50.44$\pm$0.68 & 50.50$\pm$0.78 & 80.74$\pm$0.47 & 78.61$\pm$0.31 & 79.51$\pm$0.28 & 79.29$\pm$0.34 & 82.08$\pm$0.46 & {\bf82.94$\pm$0.13} \\
        & $n=100$ & 49.04$\pm$0.80 & 42.11$\pm$0.95 & 78.89$\pm$0.90 & 71.60$\pm$1.52 & 74.62$\pm$0.72 & 73.16$\pm$0.65 & 80.22$\pm$0.24 & {\bf81.41$\pm$0.35} \\
        \midrule
        \multirow{2}{*}{CIFAR100} 
        & $n=20$ & 32.24$\pm$0.32 & 32.50$\pm$0.55 & 52.12$\pm$0.22 & 40.71$\pm$0.05 & 38.42$\pm$0.71 & 44.67$\pm$0.37  & 55.27$\pm$0.30 & {\bf57.25$\pm$0.07} \\
        & $n=100$ & 29.13$\pm$0.18 & 14.27$\pm$0.35 & 44.45$\pm$0.64 & 21.61$\pm$0.21 & 27.07$\pm$0.25 & 26.48$\pm$0.27 & 48.91$\pm$0.77 & {\bf50.20$\pm$0.39} \\
        \bottomrule
        \end{tabular}}
\end{table*}

\begin{table*}[!htb]
    \centering
    \caption{Test accuracy~(\%) on different datasets in the pathological setting}
    \label{tab:main_pathological}
    \resizebox{1.0\linewidth}{!}{
        \begin{tabular}{llcccccccc}
        \toprule
        Dataset & Client & FedAvg & FedProx & FedPer & LG-FedAvg & pFedMe & FedFomo & pFedSD & FedCKD  \\ 
        \midrule
        \multirow{2}{*}{FMNIST} 
        & $n=20$ & 75.71$\pm$0.27 & 75.40$\pm$1.06 & 99.42$\pm$0.01 & 99.15$\pm$0.04 & 98.76$\pm$0.03 & 99.28$\pm$0.01 & 99.45$\pm$0.01 & {\bf99.49$\pm$0.01} \\
        & $n=100$ & 84.13$\pm$0.44 & 74.38$\pm$0.90 & 97.28$\pm$0.12 & 95.70$\pm$0.08 & 94.87$\pm$0.03 & 96.14$\pm$0.41 & 97.42$\pm$0.04 & {\bf97.49$\pm$0.02} \\
        \midrule
        \multirow{2}{*}{CIFAR10} 
        & $n=20$ & 45.37$\pm$0.53 & 45.86$\pm$1.12 & 91.69$\pm$0.10 & 92.21$\pm$0.36 & 89.69$\pm$0.39 & 91.66$\pm$0.17 & 92.52$\pm$0.24 & {\bf92.83$\pm$0.16} \\
        & $n=100$ & 43.52$\pm$1.70 & 38.24$\pm$1.39 & 86.66$\pm$0.24 & 79.25$\pm$1.63 & 80.47$\pm$1.10 & 81.62$\pm$0.54 & 86.81$\pm$0.32 & {\bf87.24$\pm$0.20} \\
        \midrule
        \multirow{2}{*}{CIFAR100} 
        & $n=20$ & 31.92$\pm$0.23 & 32.16$\pm$0.32 & 55.57$\pm$0.26 & 45.48$\pm$0.63 & 38.66$\pm$1.21 & 48.84$\pm$0.11 & 59.54$\pm$0.01 & {\bf60.70$\pm$0.02} \\
        & $n=100$ & 27.88$\pm$0.35 & 13.32$\pm$0.37 & 47.01$\pm$0.50 & 18.96$\pm$0.30 & 26.96$\pm$0.30 & 25.21$\pm$0.63 & 50.99$\pm$0.43 & {\bf52.48$\pm$0.15} \\
        \bottomrule
        \end{tabular}}
\end{table*}

\myPara{Individual personalization.}~Individual personalization is critical in PFL because it allows the model to take into account each client's unique data characteristics and distribution. Fig.~\ref{fig:accuracy_difference} shows the difference in performance between our method FedCKD and the baseline pFedSD in terms of individual accuracy for each client. In the practical setting, FedCKD demonstrates superior accuracy compared to pFedSD in $17$ clients while showing lower accuracy in $3$ clients, indicating that FedCKD outperforms pFedSD in $85.00$\% of the cases. In the pathological setting, FedCKD's accuracy exceeds that of pFedSD in $14$ clients and lags behind in $6$ clients, resulting in higher accuracy in $70.00\%$ of the clients when compared to pFedSD. The findings suggest that FedCKD exhibits strong personalized performance, showcasing its robust capability in adapting to varying client data distributions.

\myPara{Individual fairness.}~The main objectives of individual fairness are to achieve high average accuracy and ensure an even accuracy distribution among clients, and the metric for individual fairness is the standard deviation of model accuracy across all clients. In our experiments in the practical setting with $100$ clients, we recorded their model accuracy and computed the standard deviation. As shown in Table~\ref{tab:main_fairness}, Our method, FedCKD, has higher test accuracy and lower standard deviation in most cases compared to other PFL methods. Notably, on the CIFAR10 dataset, FedCKD attains the highest average accuracy of $81.02\%$ and the lowest standard deviation of $8.84\%$. Moreover, on the FMNIST dataset, FedCKD attains a standard deviation similar to pFedSD. Despite a slight increase in standard deviation on the CIFAR100 dataset, FedCKD significantly surpasses other PFL methods in terms of average accuracy. Overall, our method showcases improved fairness, leading to enhanced performance across all clients.

\begin{table}[t]
  \centering
  \caption{Test accuracy~(\%) across all clients}
  \resizebox{!}{!}{
    \begin{tabular}{lccc}
    \toprule
    Method & FMNIST & CIFAR10 & CIFAR100 \\
    \midrule
    FedPer & 94.69$\pm$5.08 & 79.25$\pm$10.20 & 44.45$\pm$5.93 \\
    LG-FedAvg & 91.31$\pm$7.20 & 72.56$\pm$12.26 & 21.78$\pm$4.95 \\
    pFedMe & 92.14$\pm$6.82 & 75.43$\pm$11.29 & 26.82$\pm$5.07 \\
    FedFomo & 91.96$\pm$6.79 & 73.09$\pm$11.91 & 26.43$\pm$5.52 \\
    pFedSD & 95.92$\pm$3.88 & 80.17$\pm$9.23 & 48.62$\pm$6.00 \\
    \midrule
    FedCKD & {\bf 96.03$\pm$3.98} & {\bf 81.02$\pm$8.84} & {\bf 49.86$\pm$5.56} \\
    \bottomrule
    \end{tabular}}
  \label{tab:main_fairness}
\end{table}

\subsection{Ablation Study}

\myPara{Contribution of annealing mechanism.}~We investigate the contribution of the annealing mechanism employed in KD. The results on the CIFAR100 dataset among $20$ clients in the practical setting are illustrated in Table~\ref{tab:annealing_mechanism}. Upon implementing the annealing mechanism, the mean accuracy increases from $57.08\%$ to $57.25\%$, while the standard deviation decreases from $0.15\%$ to $0.07\%$. These results demonstrate that the annealing mechanism not only enhances the model’s performance but also increases its stability.

\begin{table}[t]
  \centering
  \caption{Test accuracy~(\%) under annealing mechanism}
  \label{tab:annealing_mechanism}
  \resizebox{!}{!}{
    \begin{tabular}{lcccc}
    \toprule 
    Method & w/o Annealing & w/ Annealing \\
    \midrule
    FedCKD & 57.08$\pm$0.15 & {\bf 57.25$\pm$0.07}\\
    \bottomrule
    \end{tabular}}
\end{table}

\begin{table*}[t]
    \centering
    \caption{Test accuracy~(\%) under various experimental settings}
    \label{tab:sensitivity_analysis}
    \resizebox{!}{!}{
        \begin{tabular}{l|ccc|ccc|ccc}
        \toprule
        \multirow{2}{*}{Method} & \multicolumn{3}{c|}{Data Heterogeneity} & \multicolumn{3}{c|}{Participation Rate} & \multicolumn{2}{c}{Model Architecture} \\
        \cmidrule(lr){2-4}\cmidrule(lr){5-7}\cmidrule(lr){8-9}
        & $\alpha=0.01$ & $\alpha=0.10$ & $\alpha=1.00$ & $r=0.20$ & $r=0.60$ & $r=1.00$ & ResNet & MobileNet \\
        \midrule
		FedAvg & 29.02$\pm$0.43 & 31.54$\pm$0.35 & 33.78$\pm$0.33 & 28.09$\pm$0.15 & 31.54$\pm$0.35 & 32.24$\pm$0.32 & 23.57$\pm$1.72 & 31.92$\pm$1.10 \\
		FedProx & 18.19$\pm$0.44 & 21.23$\pm$0.46 & 31.42$\pm$0.43 & 16.52$\pm$0.42 & 21.23$\pm$0.46 & 32.50$\pm$0.55 & 20.81$\pm$0.66 & 31.36$\pm$0.75 \\
		FedPer & 64.47$\pm$0.43 & 52.14$\pm$0.65 & 34.12$\pm$0.27 & 53.25$\pm$0.13 & 52.14$\pm$0.65 & 52.12$\pm$0.22 & 57.20$\pm$0.68 & 59.74$\pm$0.38 \\
		LG-FedAvg & 54.27$\pm$0.64 & 37.77$\pm$0.08 & 21.36$\pm$0.58 & 29.53$\pm$0.73 & 37.77$\pm$0.08 & 40.71$\pm$0.05 & 16.02$\pm$0.26 & 18.26$\pm$1.33 \\
		pFedMe & 46.97$\pm$0.67 & 38.64$\pm$0.38 & 24.95$\pm$0.35 & 36.97$\pm$0.52 & 38.64$\pm$0.38 & 38.42$\pm$0.71 & 32.96$\pm$0.98 & 42.88$\pm$0.77 \\
		FedFomo & 58.52$\pm$0.37 & 44.14$\pm$0.21 & 29.41$\pm$0.03 & 42.42$\pm$0.29 & 44.14$\pm$0.21 & 44.67$\pm$0.37 & 45.62$\pm$0.31 & 41.87$\pm$0.34 \\
		pFedSD & 67.11$\pm$0.58 & 56.61$\pm$0.34 & 37.81$\pm$0.33 & 55.24$\pm$0.55 & 56.61$\pm$0.34 & 55.27$\pm$0.30 & 59.74$\pm$0.90 & 63.28$\pm$0.28 \\
		\midrule
		FedCKD & {\bf 67.85$\pm$0.26} & {\bf 57.11$\pm$0.27} & {\bf 40.64$\pm$0.36} & {\bf 56.81$\pm$0.28} & {\bf 57.11$\pm$0.27} & {\bf 57.25$\pm$0.07} & {\bf 60.30$\pm$0.10} & {\bf 63.45$\pm$0.27} \\
        \bottomrule
        \end{tabular}
    }
\end{table*}

\subsection{Sensitivity Analysis}
We conduct a sensitivity analysis to investigate the robustness to data heterogeneity, participation rate, and model architecture on the CIFAR100 dataset. The default hyperparameters used, unless stated otherwise, are $\alpha=0.1, n=20, r=1.0$.

\myPara{Robustness to data heterogeneity.}~Data heterogeneity is a fundamental factor that directly influences the model's learning efficiency and ultimate performance. In the experiments, we set $\alpha=\{0.01, 0.10, 1.00\}$. Lower $\alpha$ values increase data distribution heterogeneity. FedCKD consistently outperforms other methods in these settings, illustrating its robustness across different levels of heterogeneity, detailed in Table~\ref{tab:sensitivity_analysis}. Remarkably, when data heterogeneity $\alpha=1.00$, FedCKD exceeds pFedSD by $2.83\%$. However, LG-FedAvg, pFedMe, and FedFomo exhibit subpar performance compared to the traditional FL methods.

\myPara{Robustness to participation rate.}~Participation rate plays a vital role in PFL as it determines the proportion of clients contributing to the global model update. In the experiments, we set $r=\{0.20, 0.60, 1.00\}$. Table~\ref{tab:sensitivity_analysis} shows that most methods exhibit reduced performance as the participation rate decreases. This decline is due to the restricted data participation in each round, which results in the model that cannot fully utilize the overall data. Noteworthy, pFedSD’s performance declines when the participation rate reaches $1.00$, likely because the wide variances in client data distributions introduce extra noise into the model updates. Conversely, our method, FedCKD, effectively mitigates the impact of noise.

\myPara{Robustness to model architecture.}~Model architecture also plays a significant role in PFL because different model architectures may exhibit varying adaptability and learning capabilities when handling specific types of data. In the experiments, we utilize two more complex models, namely ResNet-8~\cite{he2016deep} and MobileNetV2~\cite{howard2017mobilenets}. Table~\ref{tab:sensitivity_analysis} demonstrates that our method, FedCKD, significantly outperforms the baseline methods across a variety of model architectures. This superiority is evident through improved mean accuracy and reduced standard deviation, suggesting its ability to maintain consistent and efficient performance in all model architectures.

\section{Conclusion}
This paper presents FedCKD, a comprehensive knowledge distillation method for PFL, designed to address catastrophic forgetting and achieve a balance between generalization and personalization. Utilizing multi-teacher knowledge distillation, we effectively transfer knowledge from global and historical models to the local model. The global model contains generalized knowledge, while the historical model holds personalized knowledge. Employing the distillation mechanism, we achieve a balance and integration of these two types of knowledge, significantly boosting the model's performance. We also implement an annealing mechanism to further enhance performance and stability. Extensive experiments demonstrate the superiority of FedCKD over existing methods. However, it has inherent limitations as the KD process often requires additional computational resources. Future research will concentrate on implementing more efficient KD techniques in PFL to reduce computational costs.


\bibliographystyle{IEEEtran}
\bibliography{root}

\begin{thebibliography}{10}
\providecommand{\url}[1]{#1}
\csname url@samestyle\endcsname
\providecommand{\newblock}{\relax}
\providecommand{\bibinfo}[2]{#2}
\providecommand{\BIBentrySTDinterwordspacing}{\spaceskip=0pt\relax}
\providecommand{\BIBentryALTinterwordstretchfactor}{4}
\providecommand{\BIBentryALTinterwordspacing}{\spaceskip=\fontdimen2\font plus
\BIBentryALTinterwordstretchfactor\fontdimen3\font minus \fontdimen4\font\relax}
\providecommand{\BIBforeignlanguage}[2]{{%
\expandafter\ifx\csname l@#1\endcsname\relax
\typeout{** WARNING: IEEEtran.bst: No hyphenation pattern has been}%
\typeout{** loaded for the language `#1'. Using the pattern for}%
\typeout{** the default language instead.}%
\else
\language=\csname l@#1\endcsname
\fi
#2}}
\providecommand{\BIBdecl}{\relax}
\BIBdecl

\bibitem{mcmahan2017communication}
B.~McMahan, E.~Moore, D.~Ramage \emph{et~al.}, ``Communication-efficient learning of deep networks from decentralized data,'' in \emph{Artificial Intelligence and Statistics Conference}, 2017, pp. 1273--1282.

\bibitem{rana2024role}
N.~Rana and H.~Marwaha, ``Role of federated learning in healthcare systems: A survey,'' \emph{Mathematical Foundations of Computing}, vol.~7, no.~4, pp. 459--484, 2024.

\bibitem{abadi2024starlit}
A.~Abadi, B.~Doyle, F.~Gini \emph{et~al.}, ``Starlit: Privacy-preserving federated learning to enhance financial fraud detection,'' \emph{arXiv:2401.10765}, 2024.

\bibitem{chen2023dfedsn}
Y.~Chen, L.~Liang, and W.~Gao, ``{DFedSN}: Decentralized federated learning based on heterogeneous data in social networks,'' \emph{World Wide Web}, vol.~26, no.~5, pp. 2545--2568, 2023.

\bibitem{guo2023new}
W.~Guo, Z.~Yao, Y.~Liu \emph{et~al.}, ``A new federated learning model for host intrusion detection system under non-iid data,'' in \emph{IEEE International Conference on Systems, Man, and Cybernetics}, 2023, pp. 494--500.

\bibitem{tan2021towards}
A.~Z. Tan, H.~Yu, L.~Cui \emph{et~al.}, ``Towards personalized federated learning,'' \emph{IEEE Transactions on Neural Networks and Learning Systems}, vol.~34, no.~12, pp. 9587--9603, 2021.

\bibitem{jin2023personalized}
H.~Jin, D.~Bai, D.~Yao \emph{et~al.}, ``Personalized edge intelligence via federated self-knowledge distillation,'' \emph{IEEE Transactions on Parallel and Distributed Systems}, vol.~34, no.~2, pp. 567--580, 2023.

\bibitem{yi2024pfedmoe}
L.~Yi, H.~Yu, C.~Ren \emph{et~al.}, ``{pFedMoE}: Data-level personalization with mixture of experts for model-heterogeneous personalized federated learning,'' \emph{arXiv:2402.01350}, 2024.

\bibitem{you2017learning}
S.~You, C.~Xu, C.~Xu \emph{et~al.}, ``Learning from multiple teacher networks,'' in \emph{Proceedings of the International Conference on Knowledge Discovery and Data Mining}, 2017, pp. 1285--1294.

\bibitem{li2020federated}
T.~Li, A.~K. Sahu, M.~Zaheer \emph{et~al.}, ``Federated optimization in heterogeneous networks,'' in \emph{Proceedings of Machine Learning and Systems}, 2020, pp. 429--450.

\bibitem{arivazhagan2019federated}
M.~G. Arivazhagan, V.~Aggarwal, A.~K. Singh \emph{et~al.}, ``Federated learning with personalization layers,'' \emph{arXiv:1912.00818}, 2019.

\bibitem{liang2020think}
P.~P. Liang, T.~Liu, Z.~Liu \emph{et~al.}, ``Think locally, act globally: Federated learning with local and global representations,'' \emph{arXiv:2001.01523}, 2020.

\bibitem{t2020personalized}
C.~T.~Dinh, N.~Tran, and J.~Nguyen, ``Personalized federated learning with moreau envelopes,'' in \emph{Advances in Neural Information Processing Systems}, 2020, pp. 21\,394--21\,405.

\bibitem{zhang2020personalized}
M.~Zhang, K.~Sapra, S.~Fidler \emph{et~al.}, ``Personalized federated learning with first order model optimization,'' in \emph{International Conference on Learning Representations}, 2020.

\bibitem{hinton2015distilling}
G.~Hinton, O.~Vinyals, and J.~Dean, ``Distilling the knowledge in a neural network,'' \emph{arXiv:1503.02531}, 2015.

\bibitem{qin2024knowledge}
L.~Qin, T.~Zhu, W.~Zhou \emph{et~al.}, ``Knowledge distillation in federated learning: A survey on long lasting challenges and new solutions,'' \emph{arXiv:2406.10861}, 2024.

\bibitem{xiao2017fashion}
H.~Xiao, K.~Rasul, and R.~Vollgraf, ``{Fashion-MNIST}: A novel image dataset for benchmarking machine learning algorithms,'' \emph{arXiv:1708.07747}, 2017.

\bibitem{krizhevsky2009learning}
A.~Krizhevsky and G.~Hinton, ``Learning multiple layers of features from tiny images,'' \emph{Technical Report}, 2009.

\bibitem{he2016deep}
K.~He, X.~Zhang, S.~Ren \emph{et~al.}, ``Deep residual learning for image recognition,'' in \emph{Proceedings of the IEEE/CVF Conference on Computer Vision and Pattern Recognition}, 2016, pp. 770--778.

\bibitem{howard2017mobilenets}
A.~G. Howard, M.~Zhu, B.~Chen \emph{et~al.}, ``{MobileNets}: Efficient convolutional neural networks for mobile vision applications,'' \emph{arXiv:1704.04861}, 2017.

\end{thebibliography}

\end{document}